# An Event Correlation Filtering Method for Fake News Detection

Hao Li[1], Huan Wang[1*] and Guanghua Liu[2]
superlihao98@gmail.com, hwang@mail.hzau.edu.cn, guanghualiu@hust.edu.cn
1) College of Informatics, Huazhong Agricultural University
2) Department of Computer Science and Engineering, University at Buffalo, The State University of New York

## Abstract

Nowadays, social network platforms have been the prime source for people to experience news and events due to their capacities to spread information rapidly, which inevitably provides a fertile ground for the dissemination of fake news. Thus, it is significant to detect fake news otherwise it could cause public misleading and panic. Existing deep learning models have achieved great progress to tackle the problem of fake news detection. However, training an effective deep learning model usually requires a large amount of labeled news, while it is expensive and time-consuming to provide sufficient labeled news in actual applications. To improve the detection performance of fake news, we take advantage of the event correlations of news and propose an event correlation filtering method (*ECFM*) for fake news detection, mainly consisting of the news characterizer, the pseudo label annotator, the event credibility updater, and the news entropy selector. The news characterizer is responsible for extracting textual features from news, which cooperates with the pseudo label annotator to assign pseudo labels for unlabeled news by fully exploiting the event correlations of news. In addition, the event credibility updater employs adaptive Kalman filter to weaken the credibility fluctuations of events. To further improve the detection performance, the news entropy selector automatically discovers high-quality samples from pseudo labeled news by quantifying their news entropy. Finally, *ECFM* is proposed to integrate them to detect fake news in an event correlation filtering manner. Extensive experiments prove that the explainable introduction of the event correlations of news is beneficial to improve the detection performance of fake news.

*Key words:* Event correlation, Fake news detection, Kalman filter, News entropy

# 1. Introduction

With the internet becoming an inseparable part of our daily lives, social network platforms such as Facebook and Twitter have passed traditional news presses to be the primary source for people to experience news and events at anytime and anywhere, which significantly facilitates people's daily life. Unfortunately, it also provides a fertile ground for the dissemination of fake news [1]. Fake news is usually created for malicious purposes such as obtaining economic and political benefits or deceiving the public [2,3,4], which could cause serious consequences to the society if not prevented. The influence of fake news reached its peak during the 2016 US presidential election. 20 fake election stories generated 8.71 million shares and comments on Facebook, which is far more than 20 real election stories published by 19 mainstream media with 7.36 million shares [5]. Most social users have cognitive limitations over news content, and they are difficult to distinguish between real news and fake news. According to a survey by YouGov, a public opinion survey agency, 49% of people think they can distinguish fake news. However, in fact only 4% of people can recognize fake news by headline [6]. Therefore, it is extremely significant to automatically detect fake news to mitigate their negative effects.

Thus far, various approaches have been proposed to detect fake news, including both traditional learning models [7,8] and deep learning models [9,10]. Traditional learning models typically extract features from news articles and train classifiers based on extracted features. Compared with traditional learning methods, existing deep learning models have shown better performances in fake news detection due to their superior abilities of automatic feature extraction. However, deploying such deep learning models usually requires a large amount of hand-labeled news data, i.e., verified news that are labeled as real or fake. Also, accurate labels can only be obtained when the annotators have sufficient knowledge about the hidden event information. The production of such labeled news data is expensive and time-consuming. Furthermore, the dynamic nature of news leads to the decaying quality of existing labeled news, which may become outdated quickly in updated event information. Therefore, how to exploit the hidden information behind the labeled news data is critical to improve the power of deep learning models in fake news detection.

This research aims to develop a novel solution to make the most of limited labeled news from the new perspective of the event correlations of news. In the context of social media, an event is usually an interesting topic that occurs at a specific period and investigates a discussion about associated topics by the social users, which can be represented by a set of news that discuss the same topic [11]. For example, although an event might be a topic about COVID-19 vaccines, they can run dozens of correlated news stories about this topic. Existing researches of fake news detection mainly restrict individual news as a separated event and ignore their event correlations. However, due to the dynamic generation nature of news, the labeled news and unlabeled news could have likelihoods to report same events, especially hot-spot discussion events. By incorporating news-based event information, we can reduce the duplicate detections of fake news of same events to improve the detection efficiency. On the other hand, the news-sharing information captured by inherent events can provide supplementary and explainable dimensions to train deep learning models, which has potential to improve the label accuracy for unverified news. Therefore, it is potential to explore the event correlations of news to improve the detection performance of fake news.

In this paper, we incorporate the event correlations of news into the detection framework of fake news and propose an event correlation filtering method (*ECFM*), mainly consisting of four components: the news characterizer, the pseudo label annotator, the event credibility updater, and the news entropy selector. Based on textual features extracted from news by the news characterizer, the pseudo label annotator exploits the event correlations of news to assign pseudo labels for unlabeled news. In addition, the event credibility updater employs adaptive Kalman filter to optimize the event credibility in the pseudo label annotator. Further, the news entropy selector proposes news entropy to select high-quality samples from pseudo labeled news to dynamically update the training set. The main contribution of this research can be summarized as follows.

- To the best of our knowledge, we are the first to utilize the event correlations of news in fake news detection, which employs news-based event information to promote the detection performance of fake news.
- We propose *ECFM* to automatically annotate unlabeled news, which helps enlarge the size of the training set to ensure the success of deep learning models in fake news detection.
- Our proposed *ECFM* is a general method. Its integrated news characterizer can be easily replaced by different models designed for single-modal or multi-modal feature extractions.
- Our experiments show that the introduction of the event correlations of news is beneficial to improve the detection performance of fake news, which provides an explainable factor in fake news detection.

## 2. Related Work

To the best of our knowledge, this is the first research to develop the fake news detection method from the perspective of the event correlations of news. Two main lines of existing research have been proposed to automatically detect fake news: the traditional machine learning method [12-14] and the deep learning method [16-19].

Many traditional machine learning methods have been applied into fake news detection. Ahmed et al. [12] proposed a fake news detection model using n-gram analysis and machine learning techniques. They investigated and compared two different feature extraction techniques and six different machine classification techniques. Based on contextual features, Gravanis et al. [13] tested different machine learning classifiers and analyzed the possible improvement under ensemble machine learning strategies. Besides, Conroy et al. [14] put forward a model that combined semantic analysis and machine learning. Their model used support vector machine for classification by obtaining keyword sentences and frequencies. The above traditional machine learning method mainly depends on artificial features extracted from textual information or structural information of news. However, the manual extraction process of artificial features is labor-intensive [15], which severely restricts their detection performance.

Compared with the traditional machine learning method, the deep learning method has achieved performance improvement due to their superior ability of the feature extraction. Wang et al. [16] proposed an end-to-end framework that derived event-invariant features and thus benefited the detection of fake news on newly arrived events. They learned the discriminable representations for identifying fake news, and simultaneously learned the event invariant representations by removing the event-specific features. Yang et al. [2] treated the truths of news and users as latent random variables, and exploited users' engagements on social media to identify their opinions towards whether the news is fake or real. They leveraged a Bayesian network model to capture the conditional dependencies among the truths of news, the users' opinions, and the users' credibility. In addition, Shu et al. [17] built a hierarchical propagation network from the macro-level to the micro-level of fake news and true news. They performed a comparative analysis of the propagation network features of linguistic, structural, and temporal perspectives between fake news and real news. Further, Shu et al. [18] developed a sentence-comment co-attention sub-network to exploit both news contents and user comments to jointly capture explainable check-worthy sentences and user comments for fake news detection. Besides, Arjun et al. [19] proposed deep ensemble framework that classified various deep learning models into the pre-defined fine-grained categories for fake news detection. They developed the models based on convolutional neural network (*CNN*) [20] and bi-directional long short-term memory network [21], and the representations from these two models were fed into a multi-layer perceptron model for the final news classification.

However, it is often time-consuming and expensive to verify the news by labeling them as real or fake ones [22, 23]. How to make the most of limited labeled data to exert the power of deep learning models is a challenge. Existing deep learning models ignore the event correlation information of news in fake news detection. This is a great concern to produce the duplicate detections of news of same events and further limit the power of the deep learning models. The perspective of the event correlations of news provides valuable extra information to explain the truth of news, which can be incorporated into the detection process by deep learning models to promote their detection performances. In this paper, we take advantage of the event correlations of news and propose a novel event correlation filtering method for fake news detection.

Table 1: The frequently used symbols

| Symbol | Meaning | Symbol | Meaning |
|---|---|---|---|
| $X^l$ | Labeled news set. | $Y^l$ | Verified label set of $X^l$. |
| $X^u$ | Unlabeled news set. | $\widehat{Y^u}$ | Pseudo label set of $X^u$. |
| $X^s$ | High-quality sample set. | $\widehat{Y^s}$ | Pseudo label set of $X^s$. |
| $N$ | Number of known events. | $E_j$ | The $j_{th}$ event. |
| $x^l$ | A labeled news, and $x^l \in X^l$. | $t$ | Update number. |
| $x^u$ | An unlabeled news, and $x^u \in X^u$. | $\hat{C}_j(t)$ | Corrected credibility of $E_j$ at update $t$. |
| $x^s$ | A selected high-quality sample, and $x^s \in X^s$. | $Z_j(t)$ | Observed event credibility of $E_j$ at update $t$. |
| $\hat{P}_j(t)$ | Corrected error covariance of $E_j$ at update $t$. | $C_j(t)$ | Predicted event edibility of $E_j$ at update $t$. |
| $P_\theta(x^u)$ | Descriptive news credibility of $x^u$. | $dt$ | Detection threshold. |
| $Ce(x^u)$ | Optimized news credibility of $x^u$. | $H(x^u)$ | News entropy of news $x^u$. |

## 3. Methodology

In this section, we detail our proposed *ECFM*. The overview of *ECFM* is illustrated in Section 3.1. Section 3.2 explains the feature extraction mechanism of the news characterizer. Section 3.3 demonstrates how the pseudo label annotator cooperates with news characterizer to assign pseudo labels for unlabeled news. Section 3.4 clarifies how the event credibility updater utilizes the idea of Kalman filter to update event credibility. In Section 3.5, the process of how the news entropy selector selects high-quality samples to improve detection performance is detailed. Section 3.6 clearly exhibits how these four components integrate with each other to detect fake news.

### 3.1 Overview

In this research, we propose *ECFM* to take advantage of the event correlations of news to improve the detection performance of fake news. As shown in Fig. 1, our proposed *ECFM* integrates four components: the news characterizer, the pseudo label annotator, the event credibility updater, and the news entropy selector. First of all, the news characterizer employs a word embedding layer and *CNN* to extract the textual features from news. Based on the extracted textual features, the pseudo label annotator combines event credibility to assign pseudo labels for the unlabeled news. Subsequently, the event credibility updater applies Kalman filter to update the event credibility used in pseudo label annotator, which balance the predicted event credibility and the observed event credibility to discover the corrected event credibility. To guarantee the quality of pseudo labels, we design a news entropy selector to exploit news entropy and select high-quality samples from the unlabeled news. The selected high-quality samples are combined with the original labeled samples to continually update the training set to improve the training and detection performance.

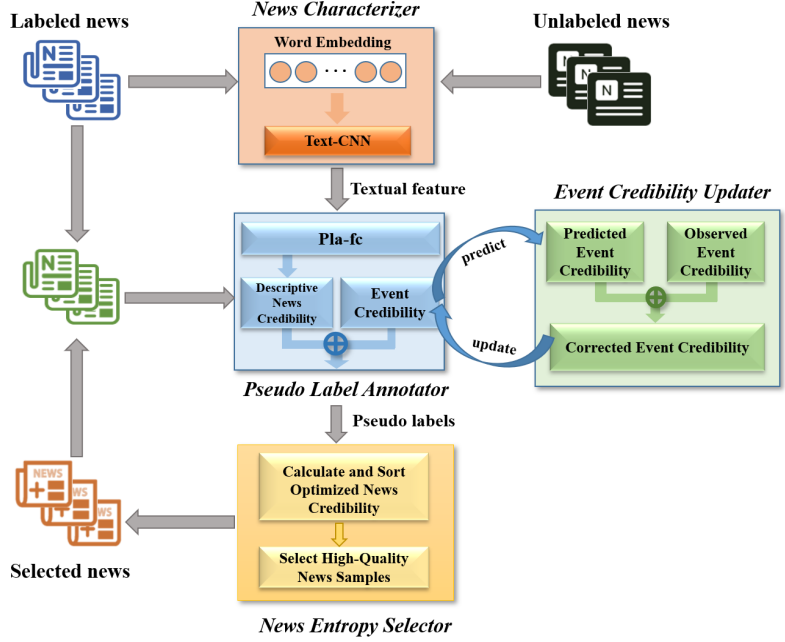

**Fig.1. The framework of *ECFM*.**

## 3.2 News Characterizer

The sequential list of words in the news is the input to the news characterizer. In order to extract textual features from the news content, we employ *CNN* as the core model of our news characterizer. *CNN* has shown its effectiveness in many fields such as computer vision [24] and text classification [25, 26]. As can be seen in Fig.2, we incorporate a modified *CNN* model, namely *Text-CNN* [27], in our news characterizer. The detailed architecture of *Text-CNN* is shown in Fig.2. As shown in Fig.2, it uses a variety of filters with different window sizes to extract textual features of different granularities.

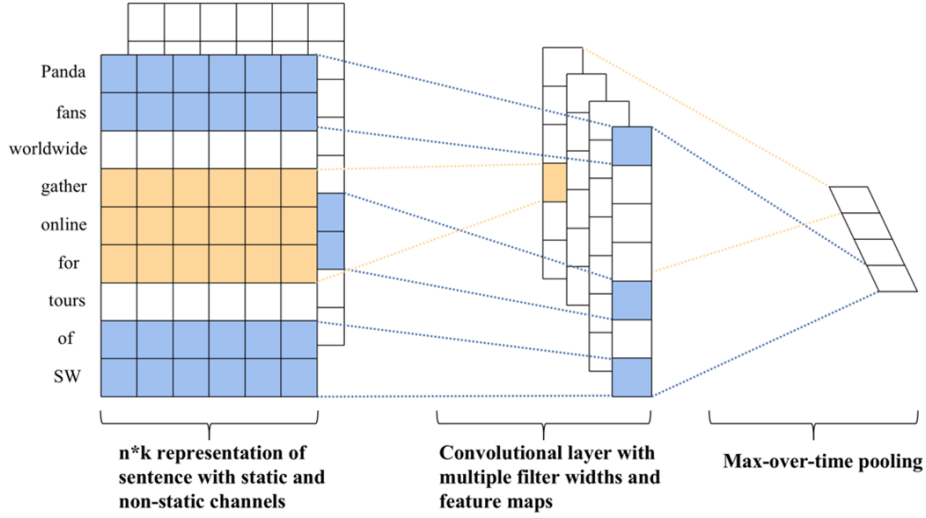

**Fig.2. The architecture of Text-CNN.**

For detailed procedures of the news characterizer, each word in the news is transformed into a word vector by a word embedding layer, namely fastText algorithm [28, 29]. For the $i_{th}$ word in the news, its corresponding $d$ dimensional word vector can be represented as $V_i \in \mathbb{R}^d$ ($1 < i < n$). Thus, the news with $n$ words can be represented as:

$$V_{1:n} = V_1 \oplus V_2 \oplus \ldots \oplus V_i \oplus \ldots \oplus V_n \tag{1}$$

Where $\oplus$ is the concatenation operator. A convolutional filter with window size $k$ selects $k$ consecutive words in the news as inputs, and one feature is outputted. In order to clearly demonstrate the process, we select $k$ consecutive words starting with the $i_{th}$ word for example, the filter operation can be represented as:

$$v_i = \sigma(W_f \cdot V_{i:i+k-1}) \tag{2}$$

Where $W_f$ represents the weight of the filter and $\sigma(\cdot)$ is the ReLU activation function. Next, we apply this filter to the rest of words of the news and get a feature vector for this news:

$$V = [v_1, v_2, \ldots, v_{n-h+1}] \tag{3}$$

For every feature vector $V$, we use max-pooling operation to select the largest feature value to extract the most important feature. Similarly, we use filters with multiple window sizes to extract textual features at different granularities. For a specific window size, we have $h$ different filters. Therefore, when there are $m$ possible window sizes, we have $m * h$ filters in total. The textual feature representation after the max-pooling operation can be written as $R_F \in \mathbb{R}^{m \cdot h}$, which is the output of the news characterizer. We denote the news characterizer as $D_c(M; \theta_c)$, where $M$, which is usually a set of textual news, is the input to the news characterizer, and $\theta_c$ represents the parameters to be learned.

### 3.3 Pseudo Label Annotator

In this section, the pseudo label annotator deploys a fully connected layer with softmax [30], denoted as Pla-fc, which incorporates the event correlations of news to predict whether the news is fake or real. The textual feature representation extracted from the news characterizer is used as the input of the pseudo label annotator.

We first train the Pla-fc by using the labeled news set $X^l$ and its verified label set $Y^l$. $x^l \in X^l$ and $y^l \in Y^l$ denote a labeled news and its verified label, respectively. We represent the Pla-fc as $D_f(\cdot; \theta_f)$, where $\theta_f$ represents all the parameters included. The descriptive news credibility of $x^l$ is defined as:

$$p_\theta(x^l) = D_f(D_c(x^l; \theta_c); \theta_f) \qquad (4)$$

Based on $X^l$ and $Y^l$, we employ cross entropy to calculate the detection loss:

$$L_f(\theta_c, \theta_f) = -E_{(x^l, y^l) \sim (X^l, Y^l)}[y^l \log(p_\theta(x^l)) + (1 - y^l)\log(1 - p_\theta(y^l))] \qquad (5)$$

The detection loss function $L_f(\theta_c, \theta_f)$ is minimized by seeking the optimal parameters $\widehat{\theta_c}$ and $\widehat{\theta_f}$. This training process is represented as:

$$(\widehat{\theta_c}, \widehat{\theta_f}) = \arg \min_{\theta_c, \theta_f} L_f(\theta_c, \theta_f) \qquad (6)$$

We further introduce the unlabeled news set $X^u$, where $x^u \in X^u$ denotes an unlabeled news. When the corresponding event of $x^u$ is the $j_{th}$ event $E_j$, the related news set of $E_j$ is represented as $\varphi(E_j)$, which contains labeled news and unlabeled news that report $E_j$. When the known credibility of the news $x' \in \varphi(E_j)$ is denoted as $nc_{x'}$, the event credibility $Ec_j$ of $E_j$ is defined as:

$$Ec_j = \sum_{x' \in \varphi(E_j)} nc_{x'} / |\varphi(E_j)| \qquad (7)$$

For the unlabeled news $x^u$, we combine $p_\theta(x^u)$ and $Ec_j$ to obtain its optimized news credibility as:

$$Ce(x^u) = \alpha \times p_\theta(x^u) + (1 - \alpha) \times Ec_j \qquad (8)$$

Here, the trade-off factor $\alpha$ is introduced to balance the weight between the descriptive news credibility and the event credibility for each unlabeled news.

To better assign pseudo labels for the unlabeled news, we calculate the optimized news credibility of each labeled news $x^l \in X^l$, and determine the detection threshold $dt$ of the optimized news credibility that can best classify the news in $X^l$ into fake news and real news based on its verified label set $Y^l$. In addition, we compare the optimized news credibility $Ce(x^u)$ of each unlabeled news $x^u \in X^u$ with the detection threshold. When $Ce(x^u) \geq dt$, $x^u$ is assigned with a pseudo real label. Else, $x^u$ is assigned with a pseudo false label. We represent the corresponding pseudo label set of $X^u$ as $\widehat{Y^u}$.

### 3.4 Event Credibility Updater

The uncertainty of the pseudo news labels in the updating process leads to the fluctuation of the event credibility in Equation (7), which further results in a variation of the optimized news credibility in Equation (8). Kalman filter is a linear, unbiased, and minimum error variance recursive algorithm to estimate the unknown state according to noisy data [31,32]. It has been widely used in many areas of industrial and government applications, such as computer vision [33], guidance and navigation systems [34], and signal processing [35]. The core of the event credibility updater is to employ adaptive Kalman filter to weaken the fluctuation in the event credibility.

Assuming the given news dataset contains $N$ known events, the state function of the predicted event credibility of $E_j$ at update $t$ is:

$$C_j(t) = \hat{C}_j(t-1) + W_j(t) \tag{9}$$

Here $\hat{C}_j(t-1)$ represents the corrected event credibility of the $j_{th}$ event $E_j$ at update $t-1$, and $W_j(t)$ is the prediction noise that follows Gauss distribution. We denote $W_j(t) \sim N_{(0,Q)}$, $Q$ is the corresponding noise covariance. At update $t$, the observed event credibility of $E_j$ is represented as:

$$Z_j(t) = Ec_j(t) + U_j(t) \tag{10}$$

$Ec_j(t)$ denotes the event credibility $Ec_j$ of $E_j$ quantified by Equation (7) at update $t$. $U_j(t)$ is the corresponding observation noise. It is similar to $W_j(t)$, where $U_j(t) \sim N_{(0,R)}$ and $R$ denotes the corresponding noise covariance. Further, we represent the corrected error covariance of $E_j$ at update $t-1$ as $\hat{P}_j(t-1)$, the prior error covariance $P_j^-(t)$ of $E_j$ at the adjacent update $t$ is predicted as:

$$P_j^-(t) = \hat{P}_j(t-1) + Q \tag{11}$$

Thus, we can get the adaptive gain $K_t$ as:

$$K_t = P_j^-(t)B^T(BP_j^-(t)B^T + R)^{-1} \tag{12}$$

Here $B$ is a transition matrix that coordinate the dimensions between $C_j(t)$ and $Z_j(t)$. The corrected credibility $\hat{C}_j(t)$ of $E_j$ is calculated as:

$$\hat{C}_j(t) = C_j(t) + K_t(Z_j(t) - BC_j(t)) \tag{13}$$

The correlated error covariance for $E_j$ is also updated as:

$$\hat{P}_j(t) = (I - K_t B)P_j^-(t) \tag{14}$$

In addition, $\hat{C}_j(t)$ and $\hat{P}_j(t)$ are further used to calculate $\hat{C}_j(t+1)$. The process of the event credibility updater is as follows.

---

**Algorithm: Event credibility updater**

INPUT: $\hat{C}_j(t-1)$ — Corrected event credibility of $E_j$ at update $t-1$

$\hat{P}_j(t-1)$ — Corrected error covariance of $E_j$ at update $t-1$

---

OUTPUT: $\hat{C}_j(t)$ — Corrected event credibility of $E_j$ at update $t$

---

1: Calculate the predicted event credibility $C_j(t)$ of $E_j$ based on $\hat{C}_j(t-1)$ according to Equation (9).

2: Obtain the observed event credibility $Z_j(t)$ based on $Ec_j(t)$ according to Equation (10).

3: Predict the prior error covariance $P_j^-(t)$ of $E_j$ based on $\hat{P}_j(t-1)$ according to Equation (11).

4: Get the adaptive gain $K_t$ according to Equation (12).

5: Find the corrected event credibility $\hat{C}_j(t)$ for $E_j$ according to Equation (13).

6: Update correlated error covariance $\hat{P}_j(t)$ according to Equation (14).

---

### 3.5 News Entropy Selector

The goal of the news entropy selector is to automatically select high-quality samples from those with pseudo labels. Inspired by information entropy [36, 37], to quantify the certainty degree of the unlabeled news $x^u \in X^u$ to own a correct pseudo label, we define the news entropy $H(x^u)$ as:

$$p'_\theta(x^u) = 1 - p_\theta(x^u) \tag{15}$$

$$H(x^u) = -\sum_{x^u \in X^u} p_\theta(x^u) \log p'_\theta(x^u) \tag{16}$$

According to the calculation process of Equation (4), $p_\theta(x^u)$ and $p'_\theta(x^u)$ denote the certainty degrees of $x^u$ to own a real label and a fake label, respectively. A larger $H(x^u)$ value has higher likelihood to own a correct pseudo label. After we obtain the news entropies of all news in $X^u$, we sort them in ascending order and select top news samples as the high-quality news. We construct the high-quality sample set $X^s$ ($X^s \subseteq X^u$) and its corresponding pseudo label set $\widehat{Y^s}$ ($\widehat{Y^s} \subseteq \widehat{Y^u}$).

### 3.5 Model Integration

In this section, we introduce how to integrate four key components: the news characterizer, the pseudo label annotator, the event credibility updater, and the news entropy selector. First, we input the labeled news set $X^l$ and unlabeled news set $X^u$ into the news characterizer to obtain their textual feature representations, denoted as $R_l$ and $R_u$, respectively. Subsequently, we train the Pla-fc based on $R_l$ and the verified label set $Y^l$ of $X^l$. Based on optimized news credibility, we assign the pseudo labels in $\widehat{Y^u}$ for the unlabeled news in $X^u$ by the pseudo label annotator. In addition, the news entropy selector selects high-quality sample set $X^s$ and their pseudo label set $\widehat{Y^s}$ from $X^u$ and $Y^u$ according to the news entropy. $X^s$ and $X^l$ are combined to update the training set. In addition, the event credibility updater adjusts the event credibility of each event in the updating process. The detection loss is calculated as:

$$L_f(X^l, Y^l, X^s, \widehat{Y^s}; \theta_c, \theta_f) = \lambda_l \cdot L_f^l(X^l, Y^l; \theta_c, \theta_f) + \lambda_s \cdot L_f^s(X^s, \widehat{Y^s}; \theta_c, \theta_f) \tag{17}$$

Here, $L_f^l(X^l, Y^l; \theta_c, \theta_f)$ and $L_f^s(X^s, \widehat{Y^s}; \theta_c, \theta_f)$ denote the detection losses on $X^l$ and $X^s$, respectively. $\lambda_l$ and $\lambda_s$ are introduced to control the balance between $L_f^l(X^l, Y^l; \theta_c, \theta_f)$ and $L_f^s(X^s, \widehat{Y^s}; \theta_c, \theta_f)$. We set the values of $\lambda_l$ and $\lambda_s$ as 1, and the two corresponding detection losses are defined by the cross entropy as:

$$L_f^l(X^l, Y^l; \theta_c, \theta_f) = -E_{(x^l, y^l) \sim (X^l, Y^l)}[y^l \log(D_f(D_c(x^l; \theta_c); \theta_f)) + (1 - y^l) \log(D_f(D_c(x^l; \theta_c); \theta_f))] \tag{18}$$

$$L_f^s(X^s, \widehat{Y^s}; \theta_c, \theta_f) = -E_{(x^s, \widehat{y^s}) \sim (X^s, \widehat{Y^s})}[\widehat{y^s} \log(D_f(D_c(x^s; \theta_c); \theta_f)) + (1 - \widehat{y^s}) \log(D_f(D_c(x^s; \theta_c); \theta_f))] \tag{19}$$

Here, $x^s \in X^S$, $\widehat{y^s} \in \widehat{Y^s}$, $x^l \in X^l$, and $y^l \in Y^l$. The detection loss function $L_f(X^l, Y^l, X^s, \widehat{Y^s}; \theta_c, \theta_f)$ is minimized by seeking the optimal parameters $\widehat{\theta_c}$ and $\widehat{\theta_f}$. This process is represented as:

$$(\widehat{\theta_c}, \widehat{\theta_f}) = arg \min_{\theta_c, \theta_f} L_f(X^l, Y^l, X^s, \widehat{Y^s}; \theta_c, \theta_f)$$

The integrated process of our proposed method is as follows.

---

**Method:** *ECFM*

---

**Input:** Labeled news set $X^l$, verified label set $Y^l$, and unlabeled news set $X^u$

---

**Output:** Pseudo label set $\widehat{Y^u}$

---

1: **for** number of training updates **do**
2:   Input the labeled news set $X^l$ and unlabeled news set $X^u$ into the news characterizer to obtain their textual feature representations $R_l$ and $R_u$.
3:   Use $R_l$ and $Y^l$ to train Pla-fc and determine the detection threshold $dt$.
4:   Use the pseudo label annotator to calculate $Ce(x^u)$ of each news $x^u \in X^u$.
5:   Assign the pseudo labels $\widehat{Y^u}$ for the unlabeled news set $X^u$ based on $Ce(x^u)$ and $dt$.
7:   Select the high-quality sample set $X^s$ and its $\widehat{Y^s}$ from $X^u$ and $\widehat{Y^u}$ by the news entropy selector.
8:   Combine $X^s$ and $X^l$ to update the training set.
9:   Use the event credibility updater to update the event credibility used in pseudo label annotator.
11: **end for**

---

## 4. Experiments

This section describes an extensive experimental study. In Section 4.1, we introduce two real-world datasets used in the experiments. In Section 4.2, we detail comparison baselines and their experimental settings. Section 4.3 clarifies the implementation detail of our proposed *ECFM*. In Section 4.4, we compare *ECFM* with five detection methods to verify its performance. As an important parameter of *ECFM*, the trade-off factor $\alpha$ balances the weight between the descriptive news credibility and the event credibility. Thus, we investigate the parameter setting of $\alpha$ in Section 4.5. In section 4.6, to illustrate the importance of the event correlations of news, we analyze the performance of our method variant. Section 4.7 conducts the pseudo label distribution analysis to analyze the training process in each update.

### 4.1 Dataset

We conduct comprehensive experiments on two real-world datasets [16, 38], which are collected from Weibo and WeChat. The details of these two datasets are as follows.

**Weibo Dataset** [39]. In this dataset, the verified fake news are crawled from May, 2012 to January, 2016 on the official rumor debunking system of Weibo. This system encourages common users to report suspicious tweets on Weibo, and a committee composed of trusted users examines the report result. According to previous work [40], this system actually serves as an authoritative source to collect fake news. The real news are collected from authoritative news sources of Xinhua News Agency. We extract the textual contents of news.

**WeChat Dataset** [38]. Its news are collected by WeChat's Official Accounts from March, 2018 to October, 2018, which are collected and sent to the experts of WeChat team for verification. The news in the training set were posted from March, 2018 to September, 2018, and the news in the testing set were posted from September, 2018 to October, 2018. There is no overlapped timestamp of news between these two sets. Note that the headlines can be seen as the summary of the news content. When preprocessing this dataset, we remove all the unlabeled news.

For these two datasets, we spilt the manually labeled news set into the training set and the testing set in a ratio 8:2. A single-pass clustering method [41] is applied to discover events from news as prior information for experiments. Thus, the event correlation of news is given, and we can obtain the labeled news and unlabeled news that report a same event. There are no overlapping events among them. The statistics of these two datasets is shown in Table 2.

Table 2: The Statistics of Datasets.

|  |  | Weibo | WeChat |
|---|---|---|---|
| Training set | Fake | 1278 | 1738 |
|  | Real | 1726 | 1264 |
| Testing set | Fake | 2398 | 2487 |
|  | Real | 3005 | 2959 |

### 4.2 Baseline Approaches

*EANN* [16]. *EANN* is one of the state-of-the-art models for fake news detection. It consists of three components: feature extractor, event discriminator and fake news detector. In this work, our input is only text. Thus, we remove the image feature extractor and keep textual extractor. The textual extractor is based on *CNN* that has the identical architecture with our *CNN* textual feature extractor. We follow the settings mentioned in [41] to cluster the headlines to get event id.

*LSTM* [21]. We use a one-layer *LSTM*. The latent representations are obtained by averaging the outputs of *RNN* [42], and these representations are fed into four fully connected layers in experiments. The hidden size of *LSTM* is set as 60. The last fully connected layer takes the 10-dimensional feature vector as input.

*CNN* [20]. The extracted textual feature in the *CNN* model is fed into one fully connect layer to adjust the dimensionality. Based on the core of *CNN*, three fully connected layers are used to predict whether this news is fake or not. We use 10 filters with the window size ranging from 2 to 5, and the fully connected layer adjusts the dimensionality of features from 40 to 30. The last fully connected layer takes 5-dimensional feature vector as input to identify whether the news is fake or not.

*LSTM$^-$*. We employ the identical *LSTM* as above in the setting of our *ECFM* method. We run the *LSTM* for 50 epochs, in each epoch, we assign pseudo labels for the unlabeled news, and select top high-quality samples from the pseudo labeled news, which are combined with the original labeled news as the training set for next epoch.

*ECFM$^-$*. It is a variant of our proposed *ECFM*, which does not consider the event correlations of news. Compared with *ECFM*, it only uses the news characterizer, the news entropy selector, and the pseudo label annotator. It is used to verify the effect of the introduction of the event correlations of news, which ignores the event correlations of news without the event credibility updater.

### 4.3 Implementation Detail

In the news characterizer, we set $k = 60$ for the dimension of the word-embedding. $h = 10$, and the window size of filters varies from 2 to 5. The size of the last fully connected layer is 10. In the Pla-fc of the pseudo label annotator, we set the hidden size of first, second, third, and fourth layer as 60, 50, 10 and 2. In the news entropy selector, if the update number is $t$, then the ratio of pseudo labeled news samples we select at this update is $2 * t\%$. In the event credibility updater, we set the initial value of the correlated error variance $P$ as 0.02, prediction noise $Q$ as 0.01, $F$ as 1.0, State transition matrix $H$ as 1.0, and observation $R$ as 0.01.

### 4.4 Performance Comparison

In the experiments, we conduct $CNN$, $LSTM$, $EANN$, $LSTM^-$ and $ECFM^-$ method as the baseline methods. To avoid the influence of randomness in the detection process, each detection method is repeated 5 independent times for both datasets. Table 3 shows the performance comparison on two real-world datasets.

On Weibo dataset, when we focus on the baselines, $ECFM^-$ outperforms other baselines with respect to the accuracy and auc-roc. However, with respect to the precision, recall and $F_1$ score, $LSTM^-$ is better than other baselines. It means we should select specific models when we concern certain evaluation metrics. Meanwhile, it also confirms the effectiveness of the selection strategy of the news entropy selector, since $LSTM^-$ and $ECFM^-$ both adopt the same selecting process of our *ECFM* method to assign labels for the unlabeled news with limited labeled news. Furthermore, we can also observe that among all the six detection methods, *ECFM* always achieves the highest detection performance in terms of all metrics. For the variant of the proposed method $ECFM^-$, it ignores the event correlations of news, and thus tends to learn the news specific features. This would lead to the failure of learning comprehensive news features among events. In contrast, with the help of the kalman filter and event correlations of news, the complete *ECFM* improves the detection performance in terms of all metrics. This demonstrates the effectiveness of the incorporating event correlations of news for performance improvements. Besides, the detection threshold is found based on the accuracy value in *ECFM*. Thus, *ECFM* is reasonable to effectively detect fake news and obtain the largest accuracy value.

Table 3: The performance comparison of different methods on two datasets

| Dataset | Method | Accuracy | AUC-ROC | Precision | Recall | $F_1$ |
|---|---|---|---|---|---|---|
| Weibo | LSTM | 0.5663 | 0.5554 | 0.5128 | 0.4579 | 0.4838 |
| | CNN | 0.5936 | 0.5590 | 0.6004 | 0.2519 | 0.3549 |
| | EANN | 0.6034 | 0.5911 | 0.5620 | 0.4821 | 0.5190 |
| | $LSTM^-$ | 0.6291 | 0.5866 | 0.7261 | 0.5081 | 0.5324 |
| | $ECFM^-$ | 0.6472 | 0.6202 | 0.6852 | 0.3795 | 0.4885 |
| | ECFM | **0.7668** | **0.7520** | **0.8099** | **0.6401** | **0.7024** |
| WeChat | LSTM | 0.5499 | 0.5552 | 0.5059 | 0.6160 | 0.5556 |
| | CNN | 0.5819 | 0.5880 | 0.5343 | 0.6582 | 0.5898 |
| | EANN | 0.6641 | 0.6682 | 0.6133 | 0.7057 | 0.6606 |
| | $LSTM^-$ | 0.5880 | 0.5951 | 0.5386 | 0.6775 | 0.6003 |
| | $ECFM^-$ | 0.5893 | 0.5910 | 0.5288 | 0.7076 | 0.6114 |
| | ECFM | **0.7936** | **0.7876** | **0.8088** | **0.7177** | **0.7605** |

On WeChat dataset, similar results can be observed as those on Weibo dataset. $LSTM^-$ still outperforms *LSTM*, which indicates the effectiveness of our selection strategy. However, *EANN* obtains better results than other baselines in terms of accuracy, auc-roc and precision, while $ECFM^-$ achieves the best performance on recall and $F_1$ score. The reason is that the data distribution of training data is different from that of the testing data, since we use the leave-one-out policy to complete the validation, which breaks the same distribution shared between the training data and the testing data on WeChat dataset. Thus, our selection strategy does not perform as well as that in the Weibo dataset. Nevertheless, our proposed *ECFM* can still achieve significant performance improvement compared to all the baselines, which is another strong prove of the effectiveness of incorporating the event correlations of news for fake news detection.

As can be seen, for the proposed *ECFM*, it outperforms all the approaches on all metrics, which shows the success of *ECFM* for fake news detection. Compared with $ECFM^-$, we can conclude that using the event correlations of news indeed improves the performance of fake news detection.

## 4.5 Parameter Analysis

In this section, we set different values of trade-off factor α in Equation (8) and analyze their performance. For each value of α, we run the model 5 times to average the metrics value. The trade-off factor α balances the weights between descriptive news credibility and event credibility. Smaller value of α means a more important role the event credibility plays in the fake news detection. When $\alpha < 0.5$, the event correlations of news plays the leading role in fake news detection, and when $\alpha > 0.5$, the information that news itself provides contributes most to detect fake news. Fig.3 presents the performance of *ECFM* under different parameter α on both datasets.

It can be seen from Fig.3 that the performance of our proposed *ECFM* is better overall when $\alpha > 0.5$ on both datasets, which means it's the descriptive news credibility not event credibility that should play the leading role in our fake news detection model. The event correlations of news could provide collective information and give an auxiliary support to predict the labels of individual news, but its importance should never exceed that of the news itself. Besides, on both datasets, our model obtains the worst results when $\alpha = 0.1$, where event correlations of news plays a dominate role to detect fake news. This shows that the overemphasis on the event correlations of news can harm the detection performance of our model.

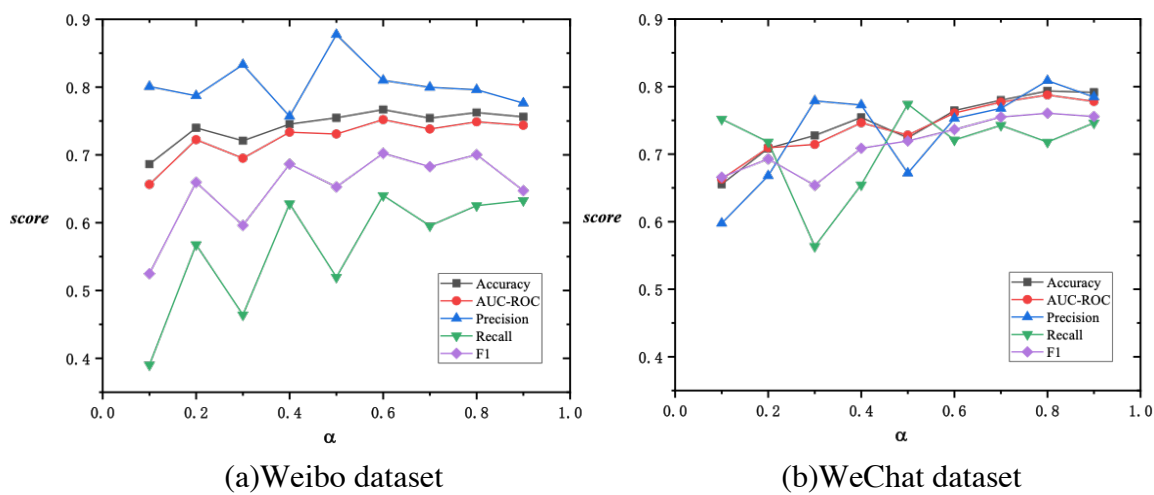

(a)Weibo dataset    (b)WeChat dataset

**Fig.3. The performance comparation under different parameter $\alpha$**

On Weibo dataset, our proposed *ECFM* method achieves best performance in terms of accuracy, auc-roc, recall and $F_1$ score when $\alpha = 0.6$, and gets the best performance on precision when $\alpha = 0.5$. As described in section 3.3, we find the detection threshold based on the accuracy value. Thus, accuracy is the most important metric to evaluate the performance of our model. Therefore, in order to achieve the best performance on Weibo dataset, we should let $\alpha = 0.6$, where the significance of the event correlations of news should approach that of news, but news should still play the leading role in fake news detection. In addition, the best accuracy score obtained is 76.68% when $\alpha = 0.6$, which increases 8% compared with the worst accuracy value obtained when $\alpha = 0.1$. The improvement is achieved with more importance put ono news and less importance put on the event correlations of news. This indicates that event correlations of news only assist the model to detect fake news, what counts most in the fake news detection task is the news itself.

On WeChat dataset, we can see the *ECFM* obtains the best performance when $\alpha = 0.8$ on accuracy, auc-roc, precision and $F_1$ score, and gets the best result on recall score when $\alpha = 0.5$. Thus, for best detection results on WeChat dataset, we let $\alpha = 0.8$, where news plays a dominate role in our fake news detection model, and the event correlations of news only plays a minor role. Furthermore, we can see that on WeChat dataset, almost a positive correlation is presented between the accuracy score and $\alpha$, which means that on WeChat dataset, the more significant role news plays in our model, the better performance could be achieved by our model. Specifically, the best accuracy score obtained by our model is 79.36% when $\alpha = 0.8$, which improves 13.8% compared with the worst accuracy score obtained when $\alpha = 0.1$. This also illustrates the superiority of news itself over the event correlations of news to predict the labels of individual news.

## 4.6 Importance of Event Correlation of News

To further demonstrate the importance of the introduction of the event correlations of news, we run the variant model $ECFM^-$ 5 times and compare the results with the results obtained by $ECFM$ stated in section 4.4. Since $ECFM^-$ also makes use of labeled news set to find the threshold value, accuracy score serves as the most significant metric to evaluate the performance of $ECFM$ and $ECFM^-$. Fig.4. shows the performance comparison of $ECFM$ and $ECFM^-$ in terms of accuracy score on both datasets. The solid line represents the average accuracy value, and the line with red color represents the performance of $ECFM$, while the line with blue color represents the performance of $ECFM^-$. As the event correlations of news can provide collective information and supplementary dimensions to train deep learning models, we can see that whatever value $\alpha$ is, $ECFM$ always achieves much better performance over $ECFM^-$ on both datasets. This demonstrates that no matter how important role event correlation plays in the fake news detection, it always benefits the detection performance which is a strong evidence for the effectiveness of the introduction of the event correlations of news.

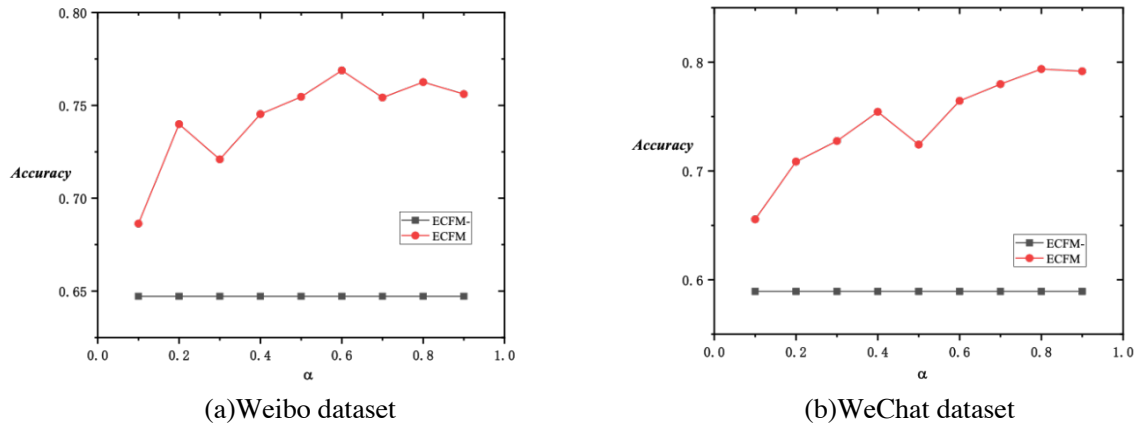

(a)Weibo dataset　　　　　　　　　　(b)WeChat dataset

Fig.4. The performance comparation for $ECFM$ and $ECFM^-$ on two datasets

## 4.7 Pseudo Label Distribution Analysis

In order to judge the quality and generalizability of our model, we record the dynamic class distribution of the pseudo labels assigned to unlabeled news in each epoch of our training process as Fig.5. In Fig.5 the ordinate represents the proportion of positive samples in all pseudo labeled samples, while the abscissa represents the number of epochs. We can see that on Weibo dataset, the overall class distribution value is between 20% to 70%. This means during the training process, for every epoch, our model didn't suffer from class imbalance, which is a potential threat to cause overfitting in the training of model. In addition, as can be seen from the Fig.5, in the last 10 epochs, the class distribution value stabilizes and fluctuates between 40% to 65%. For one thing, the stabilization of the class distribution indicates the effectiveness of the introduction of kalman filter. As described in section 3.4, the kalman filter aims to weaken the variation of the optimized news credibility, since pseudo labels are assigned based on optimized news credibility and the detection threshold, the stabilization of optimized news credibility can lead to the stabilization of class distribution of pseudo labels. Thus, kalman filter can also weaken the class distribution fluctuation of pseudo labels, contributing to a more generalized model. For another, the proportion of the fake news in the testing set is about 44%, which approaches the class distribution value in the last 10 epochs. This shows model has a potential high likelihood to gain the correct pseudo labels in the last 10 epochs. On WeChat dataset, similar results can be observed. The overall class distribution value is between 25% to 60%, which also shows that our method didn't suffer from class imbalance in each training epoch and can generalize well. In addition, compared to the results of Weibo dataset, the curve of the results on WeChat dataset is more smooth, and the fluctuation is overall much weaker. In the last

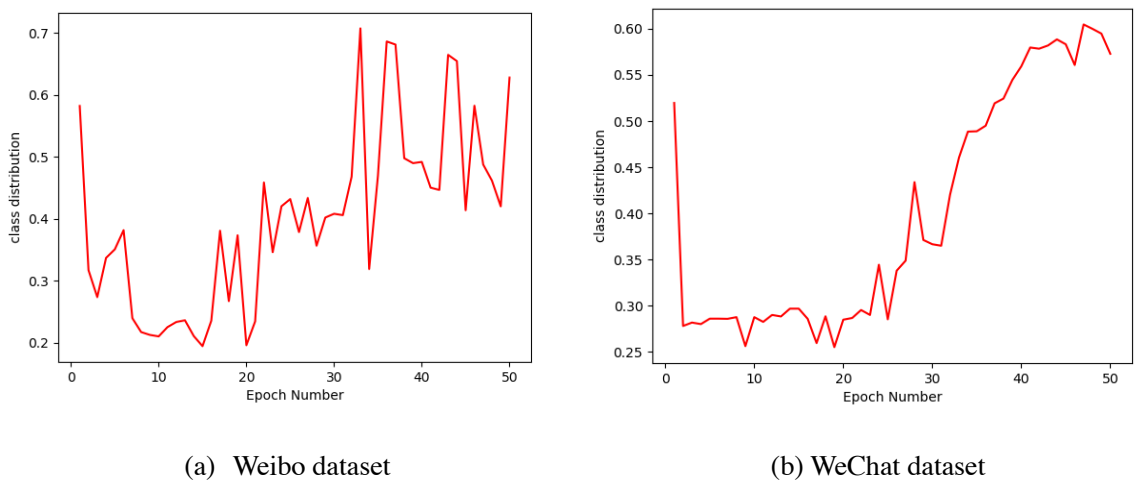

(a) Weibo dataset          (b) WeChat dataset

**Fig.5. The dynamic class distribution of pseudo labels on two datasets**

10 epochs, the class distribution value varies between 55% to 60%, while class distribution varies from 40% to 65% on Weibo dataset. Besides, the proportion of fake news in the testing set is about 46%, which approximates the class distribution value in the last 10 epochs. This again proves the effectiveness of the introduction of kalman filter and implies the high likelihood of getting the correct pseudo labels in the training process. As illustrated above, the results shown in the Fig.5. can demonstrate the high quality and generalizability of our method.

## 5. Conclusions

In this paper, we conduct a research into an important problem of fake news detection. Due to the dynamic nature of news, it's impractical to obtain continuously labeled high-quality samples to train an effective model, especially deep learning models. In order to fully unleash the power of deep learning models, it's particularly significant to make the most of the rare labeled data. Therefore, in this paper we propose a novel solution to make the most of limited labeled news from the new perspective of the event correlations of news. The proposed model mainly contains four components: the news characterizer, the pseudo label annotator, the event credibility updater, and the news entropy selector. The news characterizer cooperates with the pseudo label annotator to identify fake news. The function of event credibility updater is to optimize the event credibility used in the pseudo label annotator. After identifying fake news, the news entropy selector employs news entropy on the pseudo labeled news to choose high-quality samples to dynamically update the training set. By leveraging event correlations of news and iteratively improving the size and quality of training set, the proposed model achieves good performance on fake news detection. Extensive experiments on two real-world datasets successfully proves the effectiveness of incorporating event correlations of news for fake news detection.